# RegSpeech12: A Regional Corpus of Bengali Spontaneous Speech Across Dialects

**Md. Rezuwan Hassan[1], Azmol Hossain[4], Kanij Fatema[1], Rubayet Sabbir Faruque[1], Tanmoy Shome[6], Ruwad Naswan[2], Trina Chakraborty[3], Md Foriduzzaman Zihad[6], Tawsif Tashwar Dipto[5], Nazia Tasnim[7], Nazmuddoha Ansary[2], Md. Mehedi Hasan Shawon[1], Ahmed Imtiaz Humayun[8], Md. Golam Rabiul Alam[1], Farig Sadeque[1], Asif Sushmit[2]**

**-**

**[1] BRAC University, [2] Bangladesh University of Engineering and Technology**
**[3] Shahjalal University of Science and Technology, [4] Khulna University,**
**[5] Islamic University of Technology, [6] Daffodil International University,**
**[7] Boston University, [8] Rice University**

**-**

**Corresponding Authors:**
**Md. Rezuwan Hassan:** `md.rezuwan.hassan@g.bracu.ac.bd`
**Farig Sadeque:** `farig.sadeque@bracu.ac.bd`

---

**Abstract**

The Bengali language, spoken extensively across South Asia and among diasporic communities, exhibits considerable dialectal diversity shaped by geography, culture, and history. Phonological and pronunciation-based classifications broadly identify five principal dialect groups: Eastern Bengali, Manbhumi, Rangpuri, Varendri, and Rarhi. Within Bangladesh, further distinctions emerge through variation in vocabulary, syntax, and morphology, as observed in regions such as Chittagong, Sylhet, Rangpur, Rajshahi, Noakhali, and Barishal. Despite this linguistic richness, systematic research on the computational processing of Bengali dialects remains limited. This study seeks to document and analyze the phonetic and morphological properties of these dialects while exploring the feasibility of building computational models particularly Automatic Speech Recognition (ASR) systems tailored to regional varieties. Such efforts hold potential for applications in virtual assistants and broader language technologies, contributing to both the preservation of dialectal diversity and the advancement of inclusive digital tools for Bengali-speaking communities. The dataset[1] created for this study is released for public use.

*Keywords:* bengali dialects, low-resource language, speech recognition

## 1. Introduction

Among various means, language is primary for expressing thoughts, vital for personal and societal growth for us humans. This led computer science to advance in understanding written and spoken language, aiming to bridge gaps [33] via natural language interaction in systems. Significant advancements in this area encompass automated speech assessment for language learning [8], evaluating language disorders and therapy [25], and supporting agriculture [31]. While STT is commonly used, resources for Bangla, especially regional variations, are lacking. Extensive research such as [15] and [24] focused on modeling Bangla phonology, both theoretically and computationally but little has been explored on the diverse regional Bangla variations in Bangladesh, both in theoretical linguistics and computational speech recognition.

Despite powerful deep-learning networks, no prior research has delved into regional Bangla due to the unavailability of datasets and resources. While a speech corpus with a similar focus is known as Subak.ko [20] does exist but unfortunately, it doesn't reflect natural spoken language. A portion of the dataset samples was obtained from popular online platforms such as YouTube and Facebook. This approach reflects contemporary language usage and ensures that the dataset aligns with modern communication practices. Another dataset named OOD-Speech [27], comprising data from 17 different audio domains, is available for research purposes but even this resource lacks a regional speech domain.

The development of speech-to-text (STT) systems has outpaced the availability of resources for Bangla, particularly its regional dialects. Bangla, ranked fifth among the world's most spoken languages [3], exhibits significant regional variation. These differences, stemming from historical, geographical, and cultural processes, are reflected in lexical usage, phonetic realizations, grammatical structures, and socio-cultural expressions, reinforcing identity and communicative alignment.

[1] https://www.kaggle.com/datasets/mdrezuwanhassan/regspeech12

1. **Vocabulary:** The Standard Bangla (SB) lexicon is marked by a consistent repertoire of words, drawing from both classical literature and modern linguistic usage. In contrast, regional dialects frequently contain distinct local expressions and slang terms that are rarely encountered in formal or governmental contexts.
2. **Pronunciation:** Spoken Bangla demonstrates significant acoustic-phonetic variation across its standard and regional manifestations. This includes differences in the production of vowels and consonants, alongside specific phonetic features that delineate each dialect.
3. **Grammar:** The core grammatical architecture of Bangla is largely shared among its dialects. However, comparative analysis reveals variations in particle affixation, verbal morphology, and clause-level construction. Additionally, certain dialects may exhibit processes of grammatical attrition or novel rule formation.
4. **Cultural Influences:** Regional dialects frequently serve as linguistic repositories of localized cultural practices and traditions, integrating indigenous proverbs, idiomatic expressions, and loanwords acquired through contact with adjacent linguistic communities.
5. **Usage:** SB is employed in formal contexts such as pedagogy, broadcasting, and official correspondence. Conversely, regional dialects are predominantly utilized in informal discourse within communal settings, highlighting context-dependent linguistic adaptation.
6. **Writing and Literature:** The use of SB is a consistent feature of formal textual production, spanning newspapers, books, and academic contributions. Conversely, regional dialects are infrequently encountered in formal written contexts, being more characteristic of informal digital platforms.
7. **Geographic Variation:** The diverse dialects found in various regions of Bangladesh and India are a product of localized socio-cultural, historical, and environmental determinants.

Most existing automatic speech recognition (ASR) systems in Bangla predominantly focus on Standard Colloquial Bangla (SCB), limiting their usability for speakers of regional dialects. The modeling of regional Bangla dialects in ASR remains an open research challenge due to the scarcity of relevant datasets and linguistic resources.

Our research bridges these gaps by offering an open-sourced dataset (RegSpeech12) containing spontaneous diversified regional speech dialects.

## 2. Related Work

As the most natural and literacy-independent mode of human communication, speech provides a universal medium for interaction. Progress in speech recognition technologies offers significant potential to democratize global access to digital resources and services. These advances facilitate equitable participation by enabling users of diverse literacy levels and linguistic competencies to interact with digital platforms. Moreover, speech recognition research has expanded access to previously unavailable technologies, improved task efficiency, and enhanced user experiences, collectively reducing digital inequality and promoting inclusivity [28].

### *2.1. International speech corpora*

Although ASR systems have achieved high accuracy for widely spoken languages and adult speech, performance remains limited for Indian regional languages such as Punjabi, Telugu, and Tamil. To address this gap, Shivaprasad et al. [29] developed a dedicated dialectal database for Telugu, thereby creating one of the first systematic resources for studying dialectal variation in speech recognition. They evaluated dialect recognition using Hidden Markov Models (HMM) and Gaussian Mixture Models (GMM), reporting that GMM-based approaches yielded superior accuracy relative to HMM.
Complementing this line of work, MengEn Zhai et al. [35] investigated the Yulin dialect in China, a severely under-resourced variety. Their study involved compiling the first speech corpus for this dialect, conducting phonological analysis, and constructing a wordlist to support recognition tasks. This resource-driven approach led to a 15.42% improvement in recognition performance over conventional methods, highlighting the critical role of corpus development in advancing ASR for low-resource dialects.

### *2.2. Bangla speech corpora*

In both Bangladesh and India, speech recognition has emerged as a domain of significant academic and industrial relevance. Over the years, a range of speech datasets has been compiled and employed for training recognition models. The corpora are listed in Table 1. The corpora were developed through meticulous annotation across different linguistic units, namely phoneme, word, utterance, and sentence. These datasets not only enable the development of robust models but also provide valuable insights into the linguistic diversity of the region. Despite these contributions, Bangla remains under-resourced, with limited publicly accessible datasets and tools available for research and development [5].

| Year | Corpus Name | Size of Dataset | No. of Speakers | Publicly Available |
|---|---|---|---|---|
| 2011 | SHRUTI [10, 11, 22] | 21.64 hours | 26 males,<br>8 females | Yes |
| 2012 | IARPA-babel103b-v0.4b [7] | 215 hours | Not known | Not Publicly Available.<br>Access per application. |
| 2014 | LDC-IL [9] | 138 hours | 240 males,<br>236 females | No |
| 2014 | TDIL [32] | 43,000 audio files | 1,000 native speakers | Not Publicly Available.<br>Available for TDIL members. |
| 2018 | Bengali Connected Word Speech Corpus [19] | 62 hours | 50 males,<br>50 females | Not known |
| 2018 | Bengali Isolated Word Speech Corpus [18] | 375 hours | 50 males,<br>50 females | Not known |
| 2018 | OpenSLR [21] | 229 hours | 323 males,<br>182 females | Publicly Available under<br>Attribution-ShareAlike<br>3.0 Unported (CC BY-SA 3.0 US) |
| 2019 | ELRA [13] | 70 hours | Not known | Not Publicly Available.<br>Available for ELRA members. |
| 2020 | Bengali Speech Corpus from<br>Publicly Available Audio & Text [2] | 960 hours | 268 males,<br>251 females | No |
| 2020 | Subak.ko [20] | 241 hours | 33 males,<br>28 females | Yes |
| 2022 | Shrutilipi [6] | 443 hours | All India<br>Radio archives | Yes |
| 2022 | Common Voice Bengali Corpus [4] | 1000 hours | All 22.1k Speakers | Yes |
| 2023 | OOD-Speech: A Large Bengali Speech Recognition<br>for Out-of-Distribution Benchmarking [27] | 2000 hours | 25k Speakers | Yes |

Table 1: Available Speech Corpus for Bengali. Stats of the datasets taken from [30] mostly.

Of the few publicly accessible resources available, the OpenSLR dataset curated by Google [21] remains the most prominent. Nonetheless, its coverage is largely confined to utterance-level speech data, limiting its utility for capturing broader linguistic variation, such as dialectal differences. While accents primarily reflect variations in pronunciation, dialects encompass more extensive differences, including vocabulary and grammatical structures. In this context, dialect-focused corpora are particularly valuable because they capture both accentual and broader dialectal variation.

For Bangla regional dialects, only the Subak.ko [20] corpus stands out for its focus on dialectal diversity. Comprising 241 hours of high-quality speech (229 hours of read speech and 12 hours of broadcast speech), the corpus was collected from 61 native speakers (33 males and 28 females) representing 8 divisions and 34 districts of Bangladesh. The read speech, recorded in standard Bangla under controlled conditions, together with samples sourced from online platforms like YouTube and Facebook, captures a wide range of accents and regional speech patterns, reflecting both traditional and contemporary language usage.

This proposed dataset was partially released through a Kaggle competition [16], in which participants attempted to develop ASR solutions to handle the linguistic diversity of Bangla dialects. The released subset, referred to here as the Ben-10 corpus, was subsequently expanded into a 78-hour linguist-validated dataset encompassing speech from ten major dialect regions across Bangladesh [12]. The extended version includes spontaneous, topic-diverse recordings and serves as a benchmark for assessing ASR model robustness under dialectal variation. The released subset constitutes the closest publicly available resource for Bangla speech covering multiple regional dialects.

## 3. Dataset

Bangla exhibits considerable linguistic diversity, encompassing numerous regional dialects that differ notably in pronunciation and accent across various regions of Bangladesh. Dialectal variation can be observed in areas such as Dhaka, Chittagong, Sylhet, Noakhali, Barishal, and Rangpur. Notably, even individual words may undergo pronunciation changes depending on the specific regional context [17].

| Subset | Bengali text | IPA Transcription |
|---|---|---|
| Standard Bengali | সামনে ইদ, ইদের কেনাকাটা করতে হবে ভাই! | ʃɐmne ɪd̪, ɪd̪er kenɐkatɐ kort̪e hɔbe bʱɐ͡ɪ! |
| Rangpur | সামনোত ইদ, ইদের কেনাকাটি কইরবার নাগবে, ভাই! | ʃɐmnot̪o ɪd̪, ɪd̪er kenɐkatɪ ko͡ɪrbar nɐgbe, bʱɐ͡ɪ! |
| Kishoreganj | সামনে ঈদ, ঈদের কিনাকাডা করন লাগবো ভাই! | ʃɐmne ɪd̪, ɪd̪er kɪnɐkɐdɐ kɔron lɐgbo bʱɐ͡ɪ! |
| Narail | ছামনে ঈদ, ঈদির কিনাকাটা ওত্তি অবে ভাই! | cʰɐmne ɪd̪, ɪd̪ɪr kɪnɐkɐtɐ ot̪t̪ɪ ɔbe bʱɐ͡ɪ! |
| Chittagong | সামনো দি ঈদ, ঈদর কিনাকাটা গরন ফরিবু বাই! | ʃɐmno d̪ɪ ɪd̪, ɪd̪or kɪnɐkɐtɐ gɔron pʰorɪbu bɐ͡i̯! |
| Narsingdi | সামনে ইদ, ইদের কিনাকাডা করুন লাগবো ভাই! | ʃɐmne ɪd̪, ɪd̪er kɪnɐkɐdɐ korun lɐgbo bʱɐ͡i̯! |
| Tangail | সামনে ইদ, ইদের কিনাকাটা করন নাগবো ভাই! | ʃɐmne ɪd̪, ɪdeɾ kenɐkɐtɐ kɔɾon nɐgbe bʱɐ͡ɪ̯ |
| Habiganj | সামনে ইদ, ইদের লাইগ্গা কেনাখাডা খরোন লাগবো বাই! | ʃɐmne ɪd̪, ɪdeɾ lɐ͡ɪggɐ kenɐkʰadɐ kʰɔɾon lɐgbo bɐ͡ɪ̯! |
| Sylhet | সামনে ইদ। ইদোর কেনাকাটা করা লাগবো, বাই। | ʃɐmne ɪd̪, ɪdoɾ kenɐkɐtɐ koɾɐ lɐgbo bɐ͡ɪ̯. |
| Barishal | হোমনে ইদ, ইদের কেনাকাডা করতে অইবে বাই। | homne ɪd̪, ɪdeɾ kenɐkɐdɐ koɾte o͡ɪbe bɐ͡ɪ̯. |
| Sandwip | সামনে ইদ, ইদের কিনাকাডা কইরতো হইব বাই। | ʃɐmne ɪd̪, ɪdeɾ kɪnɐkɐdɐ ko͡ɪɾt̪o ho͡ɪbo bɐ͡ɪ̯. |
| Cumilla | সামনেদো ইদ, ইদের কিনাকাডা করতইবো বাই। | ʃɐmned̪o ɪd̪, ɪd̪er kɪnɐkɐdɐ koɾt̪o͡ɪbe bɐ͡ɪ̯. |
| Noakhali | সামনে ঈদ আইয়্যের, ঈদের বাজার হদাই করন লাইগবো ভাই। | ʃɐmne ɪd̪ a͡ɪer, ɪdeɾ baɟɐɾ hɔdoɪ kɔɾon lɐ͡ɪgbo bʱɐ͡ɪ̯. |

Table 2: Dialect diversification from different regions

We developed a speech corpus exceeding one hundred hours, with a deliberate focus on capturing regional dialectal variation in bangla. The recordings were collected from speakers across Rangpur, Kishoreganj, Narail, Chittagong, Narsingdi, Tangail, Barishal, Habiganj, Sandwip, Sylhet, Noakhali, and Cumilla. Table 2 illustrates the linguistic diversity observed among dialects from different regions. IPA transcriptions accompany only these examples for clarity and are absent elsewhere in the dataset. Note that although Kanij et al. [14] introduced a framework for IPA transcription of Standard Colloquial Bangla; however, no analogous framework is available for regional dialects. Therefore, the IPA transcriptions presented here were manually prepared by a professional linguist following analysis of the transcriptions and discussions with native speakers.

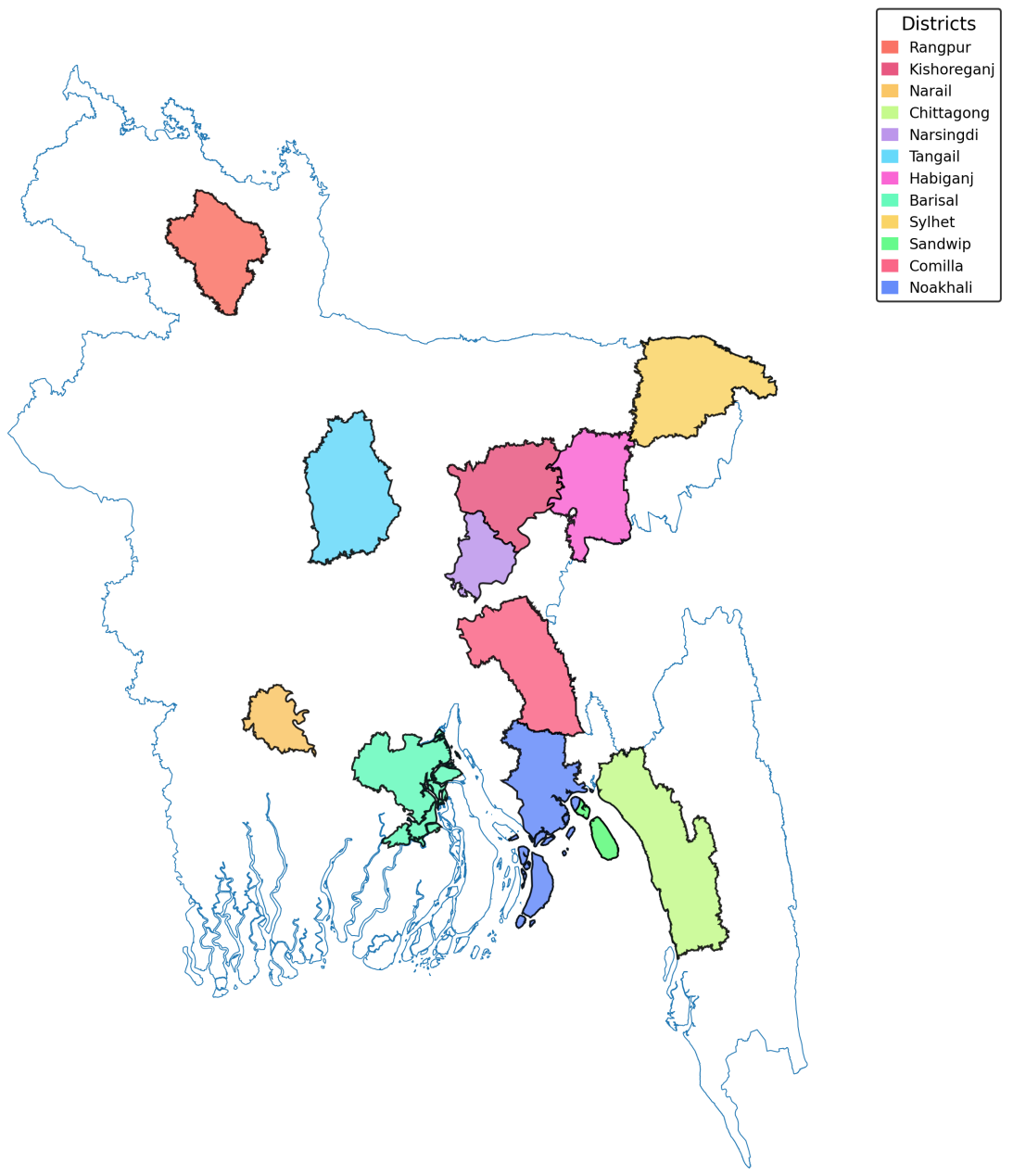

Figure 1: Mapped regions of the dialects along with reference point

This process also included carefully gathering data following specific protocols to encourage natural speech from the aforementioned regions. We obtained monologues from individuals, allowing them to express their thoughts and feelings freely.

### *3.1. **Data collection and validation***

Our objective was to develop a speech corpus that would enable any ASR model trained on it to accurately recognize bangla speech from a variety of speakers across different topics, without exhibiting bias. To achieve this goal, several features were incorporated to ensure diversification, as outlined below.

- *Voice diversification:* To ensure a representative dataset and to mitigate potential biases toward any individual speaker, we deliberately recruited a broad pool of participants from diverse backgrounds. Each recording session was standardized, producing audio clips of approximately ten minutes in duration, thereby maintaining consistency across samples.

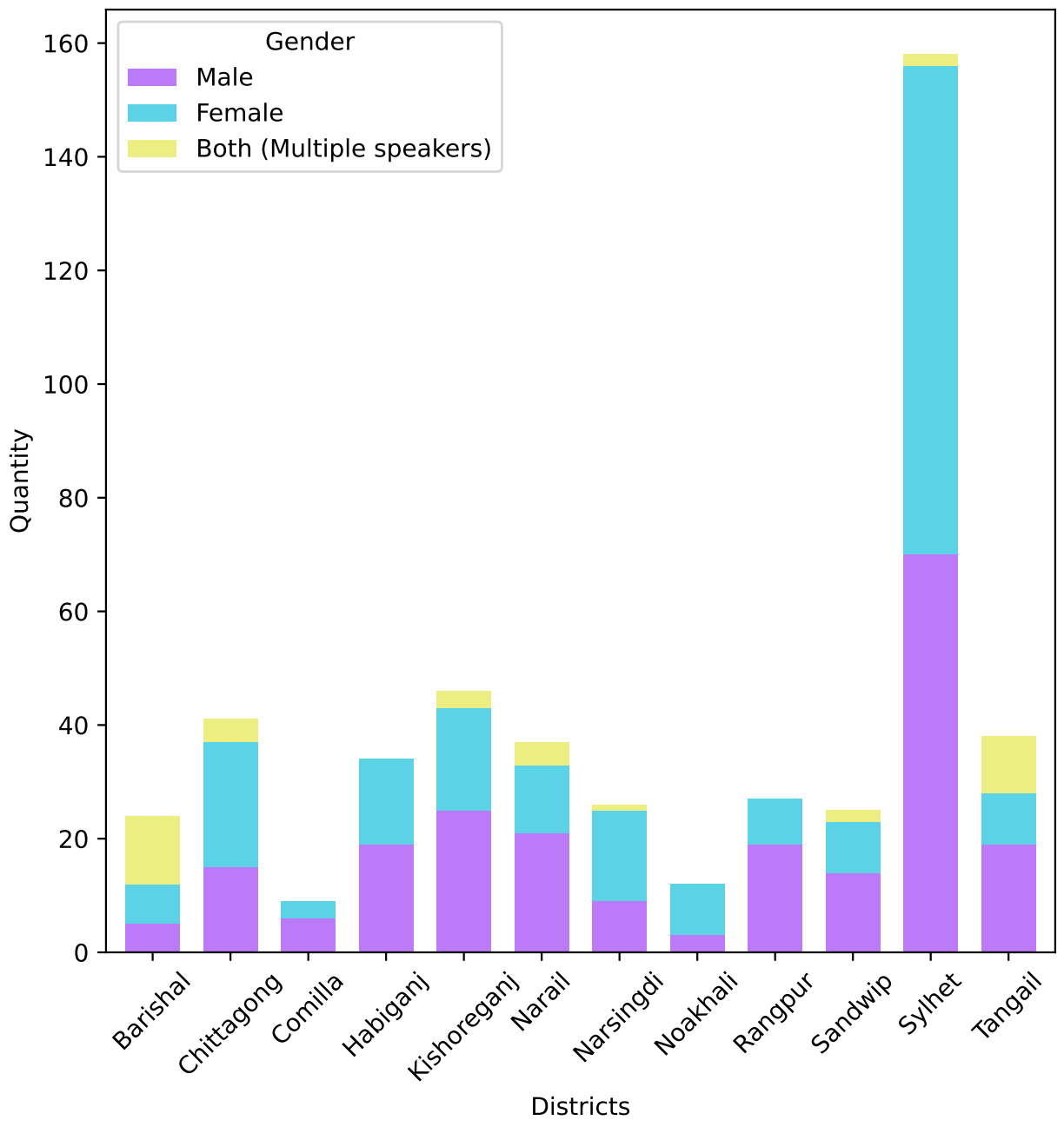

Figure 2: Gender quantity in the Regional Speech Corpus

- *Gender diversification:* In constructing the speech corpus, we placed emphasis on achieving a balanced distribution of genders, aiming for a 50:50 ratio between male and female voices. The final collection consists of contributions from at least 394 speakers. Of these, approximately $53.04\%$ are male and $37.56\%$ are female, while $9.39\%$ of the recordings include mixed-gender interactions with multiple speakers. This corresponds to $209$ male participants, $148$ female participants, and 37 clips featuring both genders. The overall gender distribution across districts is illustrated in Fig. 2.

- *Age diversification:* Age is an important factor influencing vocal characteristics, as physiological changes such as reductions in lung capacity and muscle strength contribute to observable alterations in speech production [26]. With this in mind, we intentionally incorporated speakers across a broad range of age groups into our dataset. The inclusion of such diversity ensures that the ASR system trained on this corpus is better equipped to recognize and accurately process speech from individuals at different stages of life.

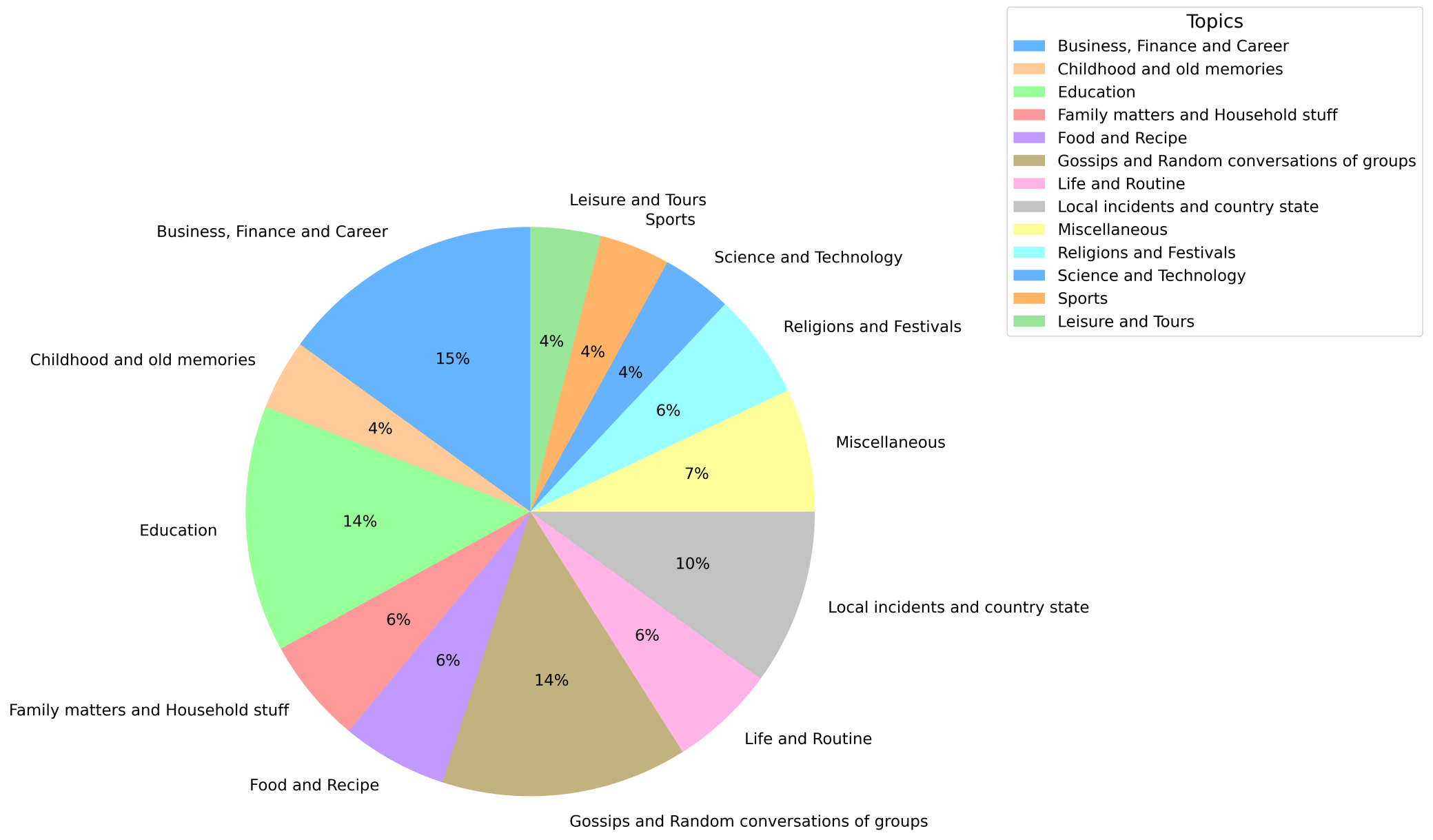

Figure 3: Distribution of corpus instances across topical categories

- *Topic diversification:* In designing the dataset, we emphasized topical diversity to enable the ASR system to generalize effectively across different domains of discourse. Data collectors were instructed to engage in everyday conversations covering a wide spectrum of subjects, including but not limited to education, family life, economics, sports, and politics. This strategy ensures that the corpus reflects the variability of real-world communication, thereby enabling the evaluation of the ASR system's ability to recognize and interpret speech across different thematic contexts. In total, the dataset comprises conversations spanning 64 distinct topics, which have been organized into 13 broad thematic categories which are illustrated in Fig. 3.

| Region | Rangpur | Kishoreganj | Narail | Chittagong | Narsingdi | Tangail | Total |
|---|---|---|---|---|---|---|---|
| **Total** | 27 | 46 | 37 | 41 | 26 | 38 | 215 |
| **Unique** | 11 | 20 | 28 | 30 | 19 | 30 | 138 |
| | | | | | | | |
| **Region** | **Barishal** | **Habiganj** | **Sandwip** | **Sylhet** | **Noakhali** | **Cumilla** | **Total** |
| **Total** | 24 | 34 | 26 | 77 | 12 | 9 | 182 |
| **Unique** | 15 | 16 | 16 | 33 | 11 | 8 | 99 |

Table 3: Total and Unique conversational topic counts across dialect regions.

Table 3 reports the distribution of conversation topics across the 12 regional subsets of the corpus. For each region, both the total number of recorded conversational instances and the number of unique topics covered are presented.

Tables 4 and 5, located in the Appendix, display representative examples of topic-specific samples obtained from the training subset. English translations are included solely for these illustrative instances and are not available across the entire dataset. The results indicate that while some regions such as Sylhet and Chittagong exhibit a wide topical diversity, others such as Cumilla and Noakhali contain fewer distinct conversational categories. Importantly, since human conversations naturally shift between subjects, a single clip may encompass multiple topics over time. To address this, each recording was manually inspected, and the duration-wise dominant topic within the conversation was assigned as the representative label. Overall, the corpus captures 237 unique topics spanning 12 regions, thereby ensuring comprehensive coverage of dialectal and thematic variation for downstream ASR benchmarking.

- *Geographical diversification:* Geographical variation plays a decisive role in shaping the dialectal diversity of bangla [34]. In order to capture such variation, our dataset was designed to include speech samples from 99 subregions across 12 regions. For example, recordings obtained from Narsingdi and Kishoreganj or Sylhet and Habiganj reveal relatively subtle phonological shifts in pronunciation, while retaining vocabulary and syntactic patterns closely aligned with standard colloquial bangla, thus maintaining high intelligibility among speakers. In contrast, the dialects of Rangpur and Chittagong present more pronounced divergences, particularly in their lexical inventories and morphological structures. Despite these differences, such variations remain authentic regional reflections of the bangla language. Table 2 presents concrete examples of these dialectal influences, underscoring the extent to which geography conditions phonological, lexical, and orthographic distinctions across regions. Table 6 in the Appendix presents a detailed list of subregions and their respective totals..

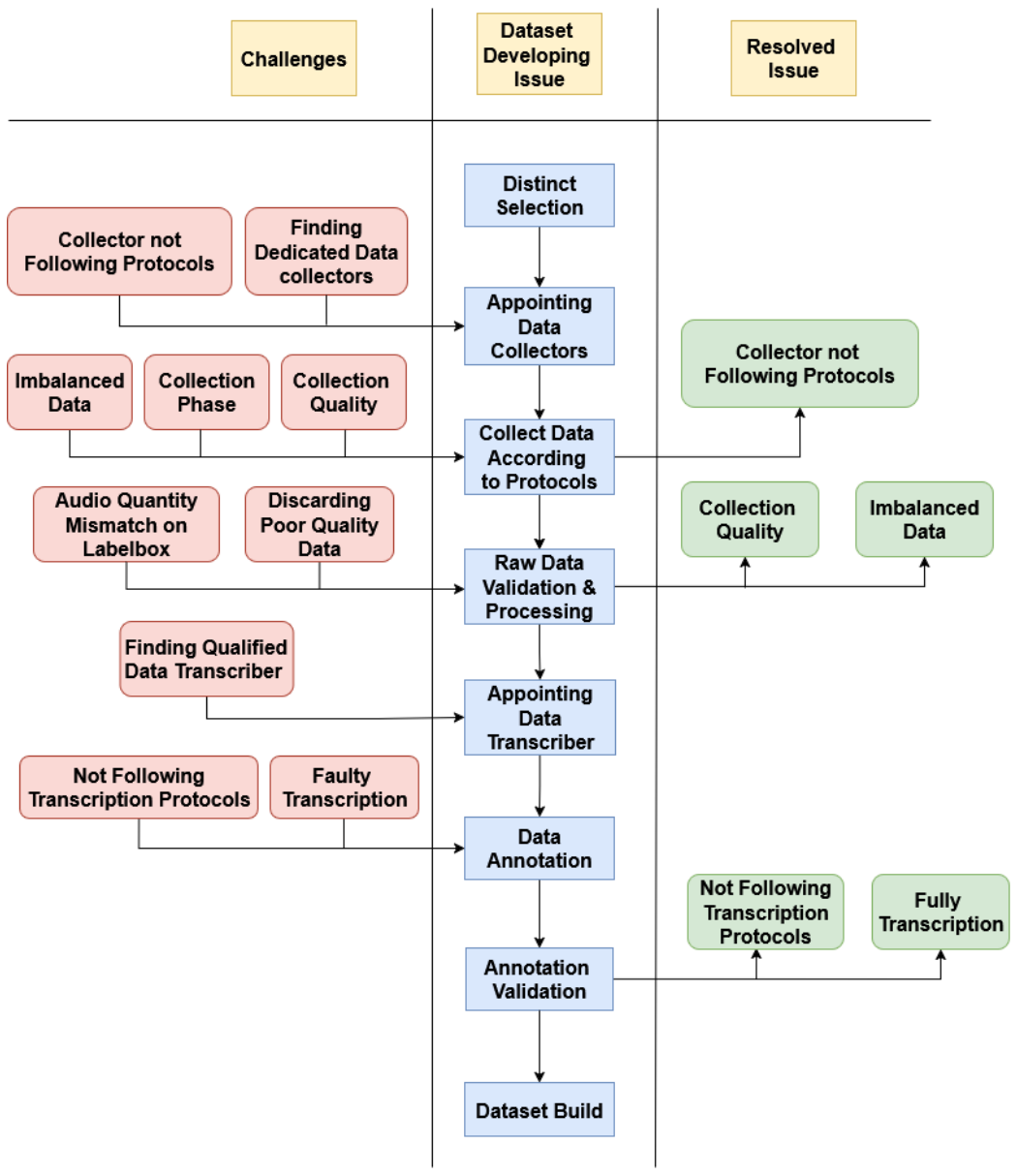

Figure 4: Top-level overview of the research

The incorporation of regional diversification features served as the foundation for our data collection strategy. The overall workflow of this process is summarized in Figure 4, while detailed explanations are presented in the subsequent discussion.

### *3.2. District Selection*

District selection for the speech corpus was guided by two principal considerations.

- Available Acquaintances: Districts were selected partly on the basis of existing personal contacts, who could contribute to data gathering efforts through direct aid or delegated work.

- Geographical Distance: To ensure diversity in language representation, districts separated by significant distances were targeted, as linguistic uniformity is often preserved only across short spatial zones.

### *3.3. Appointing Data Collectors*

The research design excluded in-person visits in order to capture unprompted, natural speech. Another factor was that, as outsiders to the community, our presence could result in participants speaking less openly or deviating from their authentic accent, which would reduce linguistic fidelity.

#### *3.3.1. Challenges*

- Finding Dedicated Data Collectors: Ensuring the five targeted dimensions of diversification required selecting a data collector capable of maintaining strict compliance with established protocols.

- Collector Adherence to Protocols: The success of the diversification strategy was dependent on the collector's commitment to established protocols. Lapses in adherence could compromise data quality, thereby affecting ASR system performance.

### *3.4. Collecting Data According to Protocols*

#### *3.4.1. Resolutions*

- Manual validation of the collected data enabled the identification and rectification of potential imbalances. Based on these findings, the data collector was directed to acquire supplementary material, thereby ensuring a more balanced corpus.

#### *3.4.2. Challenges*

- Collection Quality: Device limitations in phone recording led to audio artifacts, notably low volume levels and noise contamination.

- Collection Pace: Data collection exhibited irregularities, with productivity varying considerably from day to day.

- Imbalanced Data: Absence of continuous supervision risked leading to imbalances in corpus composition.

### *3.5. Raw Data Validation and Processing*

#### *3.5.1. Resolutions*

- Collection Quality: The audio clips were systematically reviewed to identify recordings with consistently low speaker volume and those contaminated by substantial background noise.

- Imbalanced Data: Through manual validation, we systematically identified specific imbalances within the dataset and implemented corrective strategies to address them.

#### *3.5.2. Challenges*

- Discarding Poor Quality Data: Upon detailed review of the collected recordings, we removed any samples that failed to comply with the established protocols or demonstrated substantially low quality.

- Audio Quantity Mismatch on Labelbox: Following audio segmentation with the VAD algorithm, any segments exceeding 30 seconds were further divided to ensure uniform duration. Subsequently, all samples were filtered to remove segments containing sound without speech and segments with inaudible speech. The processed audio was then uploaded as a single batch to the Labelbox annotation platform [1]. Occasionally, a software bug prevented complete batch uploads without notification; to address this, a verification script was executed to confirm successful upload of all segments.

### *3.6. Appointing Data Transcribers*

#### *3.6.1. Challenges*

- Finding Dedicated Data Transcribers: We appointed transcribers who were residents of the data collection districts, taking advantage of their prior exposure to local dialects and pronunciation patterns. Candidates were assessed with the aid of a linguist, who verified their command of bangla orthography, grammar, and regional accent accuracy. A final cohort of transcribers was selected based on these criteria to transcribe the audio recordings.

### *3.7. Data Annotation*

We utilized Labelbox for annotating the audio data. Transcribers were granted platform access and received explicit guidelines from a linguist on managing the spelling of out-of-vocabulary terms. Figure 5 illustrates the interface of the annotation platform.

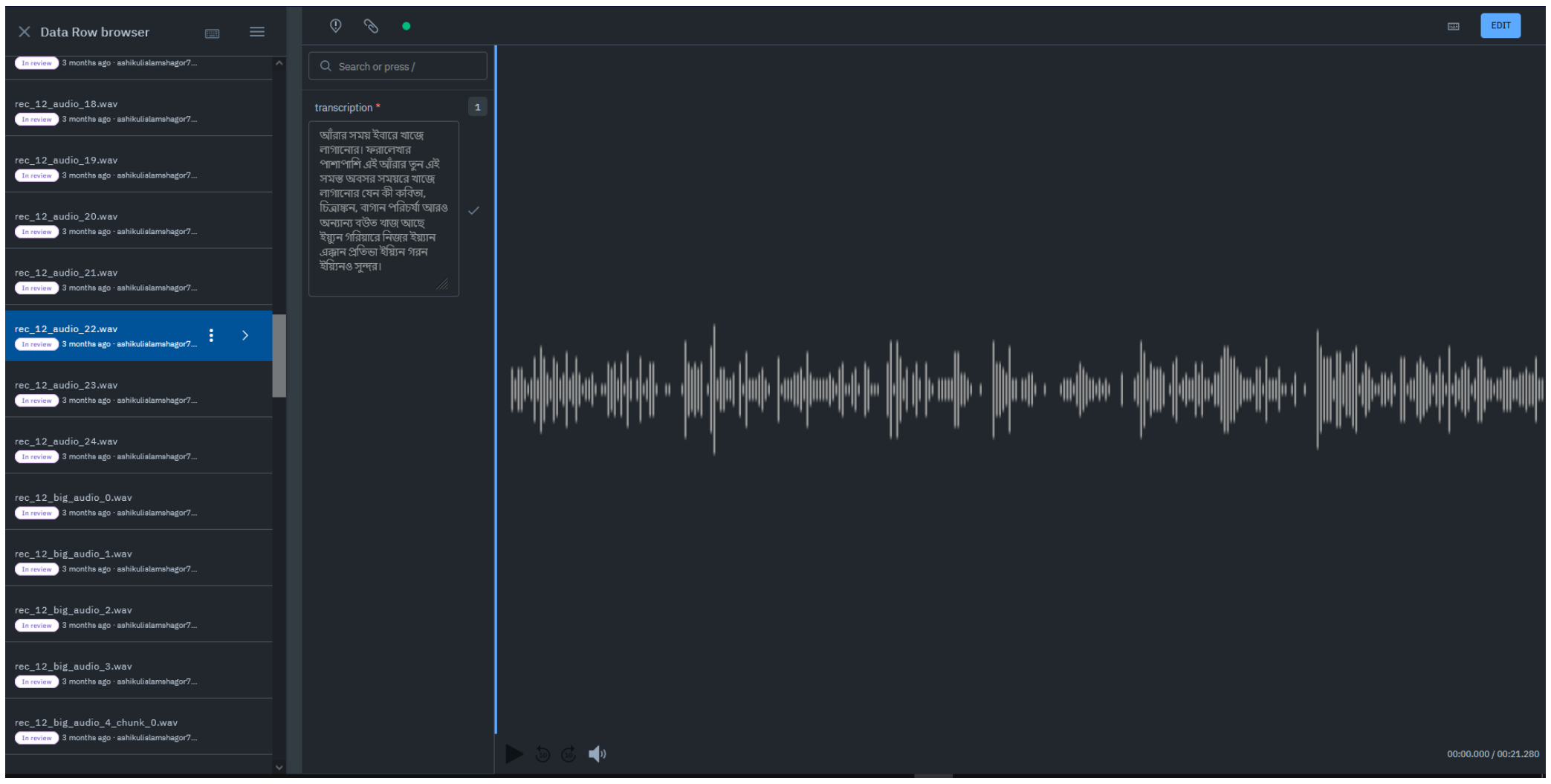

Figure 5: Interface of the annotation platform Labelbox

### 3.7.1. *Challenges*

- Faulty Transcription: Although standard guidelines are followed and instructions are clearly provided, transcription errors may still arise due to inadvertent human mistakes.

- Not Following Transcription Protocols: Failure to fully comprehend the prescribed procedures can cause certain annotators to introduce inaccuracies during transcription.

## 3.8. *Data Validation*

The annotation process involves a thorough review of spelling and transcription by a linguist, though occasional errors may be unavoidable

### 3.8.1. *Resolutions*

- Faulty Transcription: When transcription for a specific region has been completed, the results are organized into a CSV file that includes relevant details. Linguists examine this file and return their feedback, after which transcribers amend the errors detected. The dataset is finalized only once accuracy has been assured through this review cycle.

- Not Following Transcription Protocols: When transcribers encounter uncertainty or make mistakes, they resolve these issues either through consultation with linguists or by incorporating the feedback provided.

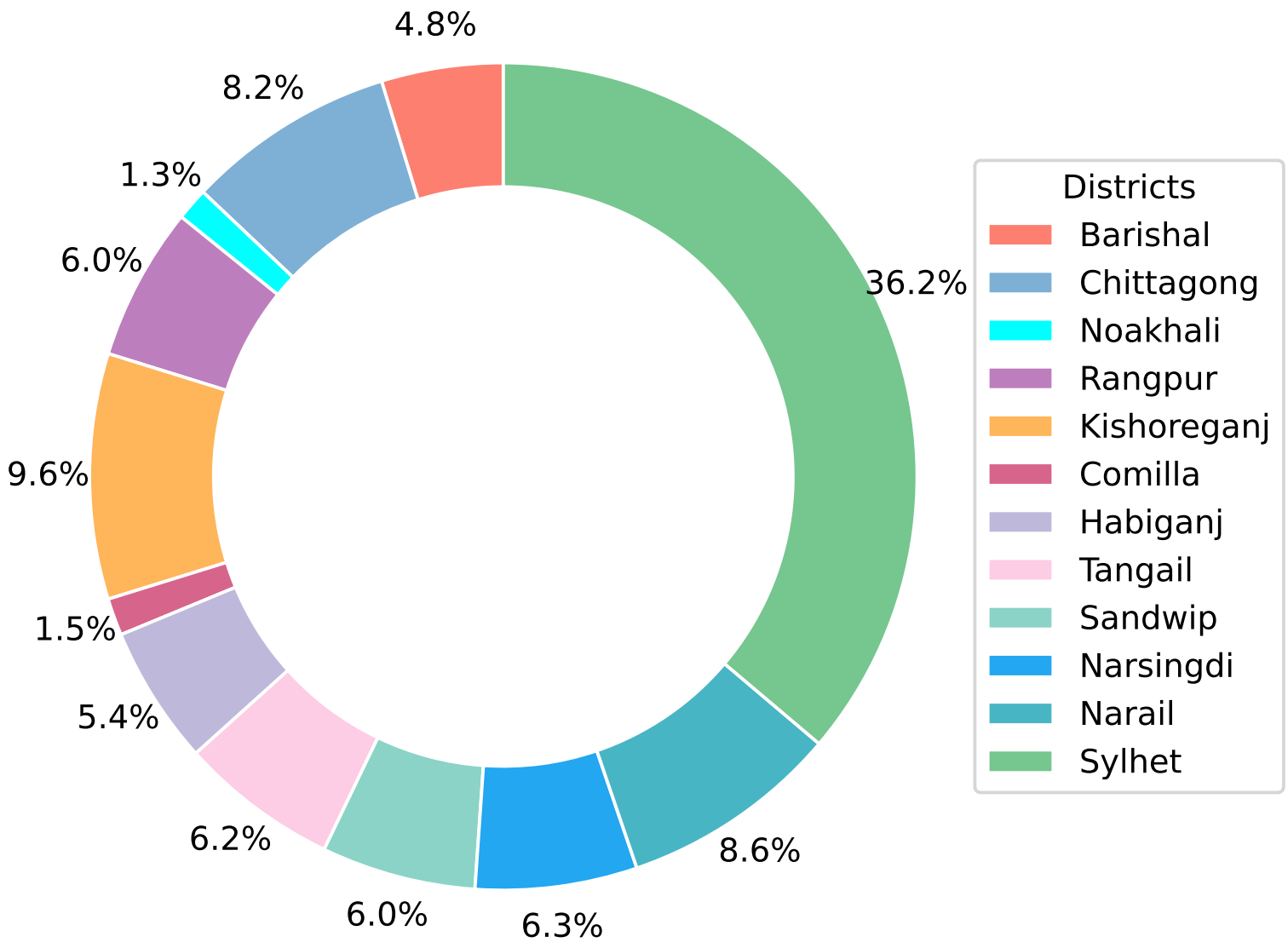

Figure 6: Percentage of each district in the corpus

### *3.9. Dataset Build*

After finalizing the transcription of all audio files, we removed incomplete or problematic entries. The curated dataset therefore contains comprehensive metadata and accurate transcriptions for each audio clip.

### *3.10. Dataset Split*

Following dataset construction, we applied an $80{:}10{:}10$ split for training, testing, and validation. The resulting partitions contained $17,049$, $2,132$, and $2,132$ samples, corresponding to approximately $80$ hours $03$ minutes, $10$ hours $06$ minutes, and $10$ hours $05$ minutes of audio, respectively.

## 4. Exploratory Data Analysis and Feature Extractions

The corpus consists of $21,313$ audio segments, amounting to more than $100$ hours of speech collected from $12$ regions. The majority of recordings are spontaneous interactions, with a smaller proportion of monologues and telephone conversations. The regional breakdown of the corpus on an hourly basis is illustrated in Figure 6.

On average, each segments span $16.94$ seconds, yielding a speech rate of $131.95$ words per minute and $37.24$ words per segment. The vocabulary includes $58,971$ distinct words, accounting for $76.05\%$ of the total lexicon. A detailed overview of the corpus is shown in the Table 7 at the appendix section.

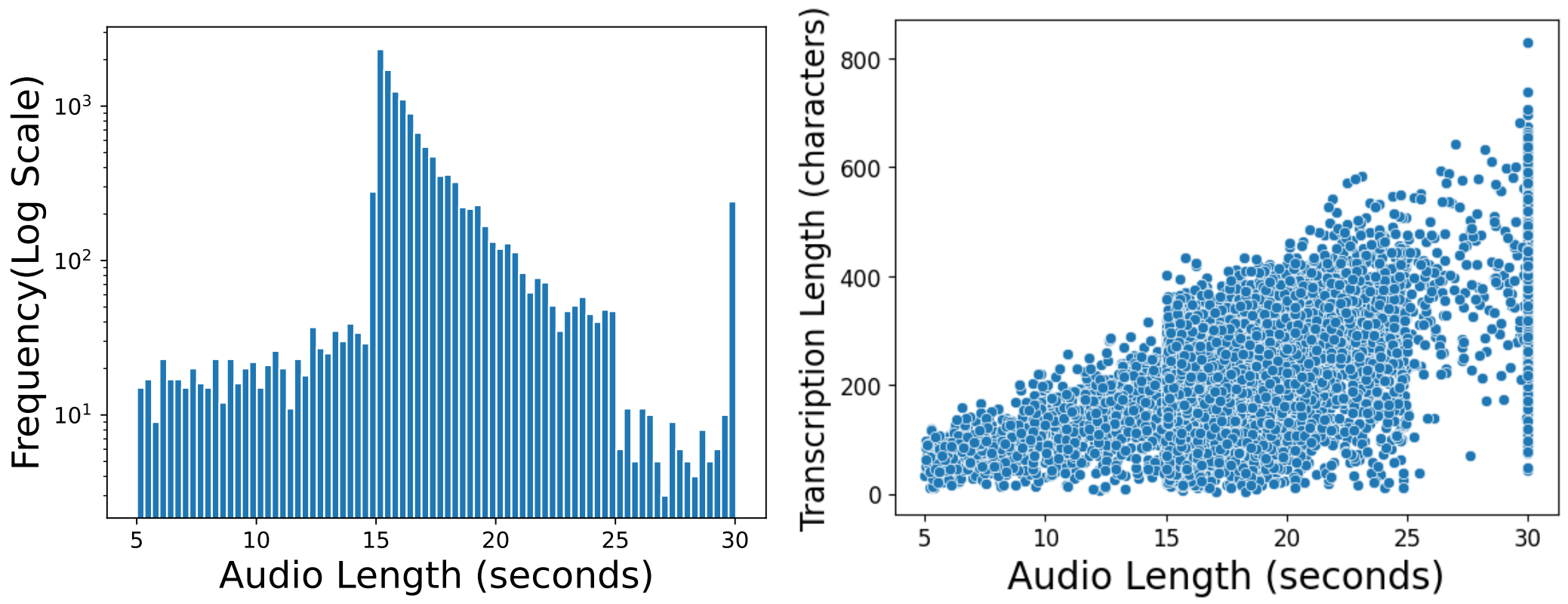

Figure 7: (Left) Audio length distribution and (Right) Scatter plot showing the relationship between Audio length and transcript length

As shown in Figure 7, most recordings are around 15 seconds long, and none surpass 30 seconds. There is also no evident correlation between the duration of a recording and the number of characters in its transcript.

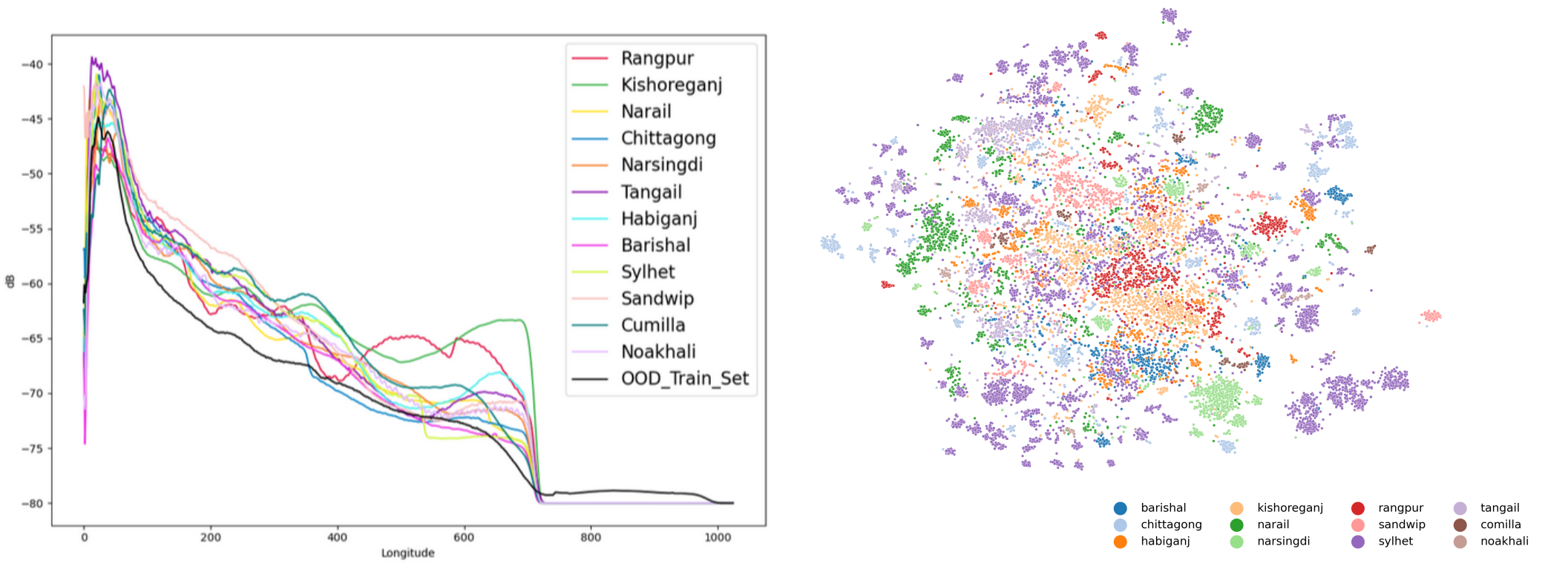

Figure 8: (Top) Long-term average spectrum of recordings representing different dialects. (Bottom) t-SNE projection of Geneva features for 12 regional subsets.

We also projected the high-dimensional Geneva feature representations into two dimensions utilizing a t-Distributed Stochastic Neighbor Embedding (t-SNE) visualization.

Figure 8(a) demonstrates the Long-Term Average Spectrum (LTAS) differences among dialects, revealing a significant distributional divergence across both semantic and spectral dimensions. Given the consistency of recording procedures, this shift is presumably a reflection of prosodic variation. Furthermore, the t-SNE analysis of 12 dialects in Figure 8(b) shows no strong intra-dialectal clustering, suggesting limited acoustic divergence and supporting the feasibility of all-purpose models. Nevertheless, lexical variation remains a critical challenge for speech recognition.

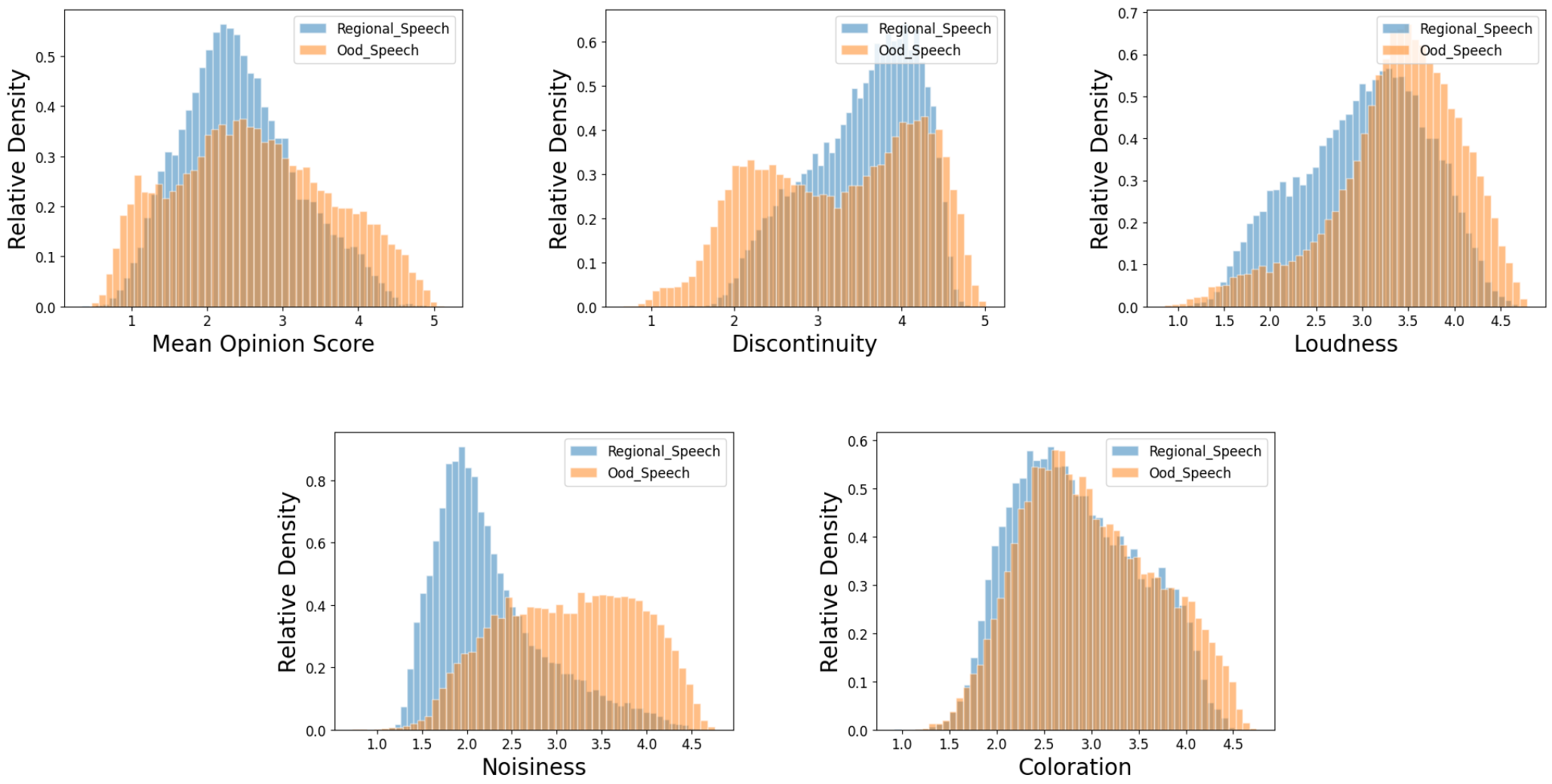

Figure 9: Comparison in distribution of NISQA features in Regional Speech (blue) and Ood Speech (orange) Datasets

Speech quality was evaluated with NISQA [23], as summarized in Figure 9. Compared with SCB OOD-Speech, the dataset shows increased noisiness due to environmental artifacts from field conditions. However, MOS and Discontinuity metrics are favorable, while Coloration values remain aligned with OOD-Speech. Loudness is diminished, reflecting limitations of phone microphones and lack of controlled mouth-to-microphone proximity. Similar Coloration metrics indicate consistent processing characteristics across both corpora.

## 5. Conclusion

The research presented in this paper presents an all-encompassing framework for constructing a regional speech corpus exceeding 100 hours from 12 regions specifically designed to address the various regional dialects of Bangladeshi Bangla. This dataset represents the only publicly available ASR (Automatic Speech Recognition) resource specifically focused on regional dialects. It offers significant potential for applications in agentic AI, meta-learning, text-to-speech and transliteration systems, and other areas where speech technology plays a pivotal role, requiring further exploration. Data collection has been completed under a uniform protocol aimed at enhancing speech-to-text models for Bengali speech with regional dialects. We highlighted the challenges and the resolutions we faced in making the speech corpus will be crucial tasks for the development of future iterations of both the dataset and the model. Future work includes linguistic analyses of dialectal variation and the standardization of transcriptions to support automatic transliteration between regional and standard Bengali.

## Acknowledgement

We acknowledge with appreciation the contributions of all individuals involved in data collection and transcription. Informed consent was obtained from each participant, and all personal information (such as usernames, email addresses, URLs, and demographic details) was excluded to maintain anonymity.

## References

[1] Labelbox. `https://labelbox.com`. Accessed: 2023-09-19.

[2] Shafayat Ahmed, Nafis Sadeq, Sudipta Saha Shubha, Md Nahidul Islam, Muhammad Abdullah Adnan, and Mohammad Zuberul Islam. 2020. Preparation of bangla speech corpus from publicly available audio & text. In *Proceedings of The 12th language resources and evaluation conference*, pages 6586–6592.

[3] Firoj Alam, Murtoza Habib, Dil Sultana, and Mumit Khan. 2010. Development of annotated bangla speech corpora.

[4] Samiul Alam, Asif Sushmit, Zaowad Abdullah, Shahrin Nakkhatra, MD. Nazmuddoha Ansary, Syed Mobassir Hossen, Sazia Morshed Mehnaz, Tahsin Reasat, and Ahmed Imtiaz Humayun. 2022. Bengali common voice speech dataset for automatic speech recognition.

[5] Abhik Bhattacharjee, Tahmid Hasan, Kazi Samin, M Sohel Rahman, Anindya Iqbal, and Rifat Shahriyar. 2021. Banglabert: Combating embedding barrier for low-resource language understanding. *arXiv preprint arXiv:2101.00204*.

[6] Kaushal Santosh Bhogale, Abhigyan Raman, Tahir Javed, Sumanth Doddapaneni, Anoop Kunchukuttan, Pratyush Kumar, and Mitesh M Khapra. 2022. Effectiveness of mining audio and text pairs from public data for improving asr systems for low-resource languages. *arXiv preprint arXiv:2208.12666*.

[7] Bills, Aric, David, Anne, Dubinski, Eyal, Fiscus, Jonathan G., Gillies, Breanna, Harper, Mary, Jarrett, Amy, Molina, María Encarnación Pérez, Ray, Jessica, Rytting, Anton, Paget, Shelley , Shen, Wade, Silber, Ronnie, Tzoukermann, Evelyne, and Wong, Jamie. 2016. Iarpa babel bengali language pack iarpa-babel103b-v0.4b.

[8] Carol A Chapelle and Erik Voss. 2016. 20 years of technology and language assessment in language learning & technology. *Language Learning & Technology*, 20(2):116–128.

[9] Narayan Choudhary and DG Rao. 2020. The ldc-il speech corpora. In *2020 23rd Conference of the Oriental COCOSDA International Committee for the Co-ordination and Standardisation of Speech Databases and Assessment Techniques (O-COCOSDA)*, pages 28–32. IEEE.

[10] Biswajit Das, Sandipan Mandal, and Pabitra Mitra. 2011. Bengali speech corpus for continuous automatic speech recognition system. In *2011 International conference on speech database and assessments (Oriental COCOSDA)*, pages 51–55. IEEE.

[11] Biswajit Das, Sandipan Mandal, and Pabitra Mitra. 2011. Shruti bengali continuous asr speech corpus.

[12] Tawsif Tashwar Dipto, Azmol Hossain, Rubayet Sabbir Faruque, Md. Rezuwan Hassan, Kanij Fatema, Tanmoy Shome, Ruwad Naswan, Md. Foriduzzaman Zihad, Mohaymen Ul Anam, Nazia Tasnim, Hasan Mahmud, Md Kamrul Hasan, Md. Mehedi Hasan Shawon, Farig Sadeque, and Tahsin Reasat. 2025. Are asr foundation models generalized enough to capture features of regional dialects for low-resource languages?

[13] ELRA. 2010. ELRA-U-S 0031: Bangla (Bengali) Speech Corpora. http://universal.elra.info/product_info.php?cPath=37_39&products_id=1669. Accessed: 2020-01-28.

[14] Kanij Fatema, Fazle Dawood Haider, Nirzona Ferdousi Turpa, Tanveer Azmal, Sourav Ahmed, Navid Hasan, Mohammad Akhlaqur Rahman, Biplab Kumar Sarkar, Afrar Jahin, Md. Rezuwan Hassan, Md Foriduzzaman Zihad, Rubayet Sabbir Faruque, Asif Sushmit, Mashrur Imtiaz, Farig Sadeque, and Syed Shahrier Rahman. 2024. Ipa transcription of bengali texts.

[15] Abdul Hai. 1964. *Dhwonibijnan O Bangla Dhwonitottwo*, 3rd edition. Bornomichil.

[16] Ahmed Imtiaz Humayun, farigys, Mohaymen Ul Anam, Rubayet Sabbir Faruque, S M Jishanul Islam, Sushmit, and Tahsin. 2024. ভাষা-বিচিত্রা: Asr for regional dialects. https://kaggle.com/competitions/ben10. Kaggle.

[17] Protiva Rani Karmaker. Dialectical and linguistic variations of bangla sounds: Phonemic analysis.

[18] Md Farukuzzaman Khan. 2018. Construction of large scale isolated word speech corpus in bangla. Global Journal of Computer Science and Technology, 18(G2):21–26.

[19] MF Khan and MA Sobhan. 2018. Creation of connected word speech corpus for bangla speech recognition systems. Asian Journal of Research in Computer Science, pages 1–6.

[20] Shafkat Kibria, Ahnaf Mozib Samin, M Humayon Kobir, M Shahidur Rahman, M Reza Selim, and M Zafar Iqbal. 2022. Bangladeshi bangla speech corpus for automatic speech recognition research. Speech Communication, 136:84–97.

[21] Oddur Kjartansson, Supheakmungkol Sarin, Knot Pipatsrisawat, Martin Jansche, and Linne Ha. 2018. Crowd-sourced speech corpora for javanese, sundanese, sinhala, nepali, and bangladeshi bengali. In Proc. The 6th Intl. Workshop on Spoken Language Technologies for Under-Resourced Languages (SLTU), pages 52–55, Gurugram, India.

[22] Sandipan Mandal, Biswajit Das, Pabitra Mitra, and Anupam Basu. 2011. Developing bengali speech corpus for phone recognizer using optimum text selection technique. In 2011 international conference on asian language processing, pages 268–271. IEEE.

[23] Gabriel Mittag, Babak Naderi, Assmaa Chehadi, and Sebastian Mo□ller. 2021. Nisqa: A deep cnn-self-attention model for multidimensional speech quality prediction with crowdsourced datasets. arXiv preprint arXiv:2104.09494.

[24] Abul Kalam Manzur Morshed. 1997. Adhunik Bhashatatwa, 2nd edition. Noya Udyog.

[25] Andreas Maier Peter Kitzing and Viveka Lyberg □hlander. 2009. Automatic speech recognition (asr) and its use as a tool for assessment or therapy of voice, speech, and language disorders. Logopedics Phoniatrics Vocology, 34(2):91–96.

[26] Melissa Petruzzello. 2022. Why does your voice change as you age?

[27] Fazle Rabbi Rakib, Souhardya Saha Dip, Samiul Alam, Nazia Tasnim, Md. Istiak Hossain Shihab, Md. Nazmuddoha Ansary, Syed Mobassir Hossen, Marsia Haque Meghla, Mamunur Mamun, Farig Sadeque, Sayma Sultana Chowdhury, Tahsin Reasat, Asif Sushmit, and Ahmed Imtiaz Humayun. 2023. Ood-speech: A large bengali speech recognition dataset for out-of-distribution benchmarking.

[28] Mohan R. Ramaswamy, Gregory Chaljub, Oliver Esch, Donald D. Fanning, and Eric vanSonnenberg. 2000. Continuous speech recognition in mr imaging reporting. American Journal of Roentgenology, 174(3):617–622. PMID: 10701598.

[29] S Shivaprasad and M Sadanandam. 2020. Identification of regional dialects of Telugu language using text independent speech processing models. International Journal of Speech Technology, 23:251–258.

[30] Sadia Sultana, M Shahidur Rahman, and M Zafar Iqbal. 2021. Recent advancement in speech recognition for bangla: A survey. Int. J. Adv. Comput. Sci. Appl, 12(3):546–552.

[31] Polisetty Swetha and Jenega Srilatha. 2022. Applications of speech recognition in the agriculture sector: A review. ECS Transactions, 107(1):19377.

[32] TDIL. 2018. TDIL: Bengali Speech Data - ASR. http://tdil-dc.in/index.php?option=com_download&task=showresourceDetails&toolid=2000&lang=en. Accessed: 2020-01-28.

[33] Mike Wald and Keith Bain. 2008. Universal access to communication and learning: the role of automatic speech recognition. Universal Access in the Information Society, 6:435–447.

[34] George Yule. 2022. The study of language. Cambridge university press.

[35] MengEn Zhai, LiHong Dong, Yi Qin, and FeiFan Yu. 2022. The research of chain model based on cnn-tdnnf in yulin dialect speech recognition. In 2022 7th International Conference on Image, Vision and Computing (ICIVC), pages 883–888.

## 6. Appendix

| Sample | Region | Topic | Bangla Transcription | English Translation |
|---|---|---|---|---|
| train_barishal_0012.wav | Barishal | Life and Routine | তবে হচ্ছে কী জানো? মানে ফ্যামিলি ছেড়ে ফ্যামিলি ছেড়ে এই ফাস্টাইম, ভার্সিটিতে যাওয়া মানি ইপতার করছি ঠিকই ভাই মানি কেমন যেনো মনের ভিতর একটা না একটা না কেমন কেমন একটা জড়তা কাজ কোরছে। মানে ভালোই লাগে না। এই মানি এতোগুলো রোজা গেলো, এই রোজা যাওয়ার পর বাসায় আসলাম একুশ রোজা | You know what's happening? I mean, after leaving my family and coming to the hostel for the first time—even though I'm going to university and doing Iftaar—I still feel a kind of emptiness inside. I don't really feel good. So many days of Ramadan have already passed, and after all those fasting days, I finally came home on the twenty-first day of Ramadan. |
| train_chittagong_0052.wav | Chittagong | Education | দেশত ফরিবি না বিদেশত ফরিবি? আঁই খইদি যেখোনো এক মিক্কে ফরিলেই অর। তই ফরে দিয়ে আঁর বড় ভাইও আঁরে খর। খরদে, বিদেশত যাইলে যাগুই এহন সুযোগ আছে ভালা। তারফরে আইএল্টস - ফায়েল্টস গইরলাম। আইএল্টস টাইয়েল্টস গরি সাই ভত্তি অইলাম। ভত্তি টত্তি অই দেহির দি | Where will you continue your studies — here or abroad? I said, Either one is fine, as long as I get to study somewhere. Then my elder brother also said, If you want to go abroad, just go. It is a good opportunity now. So I thought, let us try for the IELTS exam and all that. I got admitted, and after that, I realized that |
| train_comilla_0067.wav | Comilla | Business, Finance and Career | আলুর টিয়া, সারের ট্যায়া, কামলার ট্যায়া তিশ থাইকা ফাইতিরিশ হাজার ট্যাহা ফনজন যাইগগায়। ব্যাছার সময় দো মনো করেন বীছের দরে, আলুর দরে, কামলার দরে কোনো লাব নাই। কোইরা রাহে কি, অহন মনো করেন ইস্কিরিমো এই কেরি খেতো পানি দিতারতো না দেইক্কা এল্লাইগগা মানুষ | The cost of potatoes, fertilizer, and labor adds up to around thirty to thirty-five get to study somewhere. Then my elder brother thousand taka. When it's time to sell, you realize that after adding up the cost of seeds, potatoes, and labor, there is no profit at all. Still, we do it anyway. Now, under the scheme, people cannot use water from the Eri irrigation scheme for their paddy fields, so people |
| train_habiganj_0120.wav | Habiganj | Family matters and Household stuff | ফোলার তো একটা ববিষ্যত আছে, মেয়ের ওত্তো একটা ববিষ্যত আছে বালো একটা স্বামী দেইখ্যা বিয়া দিতাম। আইচ্ছা তে ইখানো আইয়া কিরখম ছলা-ফেরা খরো মাইনষের সাতে, ই এলাখার মানুষ কিরখম, না কিরখম? | The boy and my daughter have a future. I want a good groom for my daughter. Okay, now tell me—after coming here, how do you adapt and behave with people? What are they actually like? |
| train_narail_0046.wav | Narail | Gossips and random conversation | গেছেলো আম্মু গেছেলো, আবু যায় নেই, আম্মু গেছেলো। যাইয়ে তো আবার চইলেও আইছেলো। এহন তো বাড়িই আছে নাকি? হয়, হয়। কুরবানি পর্যন্ত তো আছো বাড়ি, না? হ, কুরবানি পর্যন্ত আছি, তুমি আসো দেহি বাড়া ইটু আড্ডা-মাড্ডা মারবানি। হ, দুলা আপুরা আসপে কুরবানিতি? | Mother has left. Father hasn't gone; only mother has. She went away and then came back again. She's at home now, isn't she? Yes, yes. You'll stay at home until Qurbani, right? Yes, I'll stay until Qurbani. You come—we'll try to catch up and hang out at home for a while. Okay! Will Dola Apu come for Qurbani Eid? |
| train_narsingdi_0060.wav | Narsingdi | Sports | ফুটবল খেলা আমার প্রিয় খেলা। এবার। জাতীয় খেলা তো। ওহ, আমনে তো আবার ব্রাজিল সাপোর্টার, আইচ্ছা এইবার বিশ্বকাপের মইদ্দে ব্রাজিল যে বাইর হইসিনগা ফরে কোন দলেরে সাপোট করছিলেন? সৌদিরে সাপোর্ট করছিলাম এবার আমি। ওহ প্রথমতে সৌদি? হ্যাঁ। কারণ আল্লায় সোয়াব দিয়া লাইবো। | Football is my favourite sport. It's a national sport here. Oh, you are a Brazil supporter. When Brazil got knocked out of the World Cup, which team did you support? I supported Saudi Arabia this time. You supported Saudi from the start? Yes, because I believe Allah will bless me for it. |

Table 4: Topic-specific examples from the corpus (1)

| Sample | Region | Topic | Bangla Transcription | English Translation |
| --- | --- | --- | --- | --- |
| train_noakhali_0130.wav | Noakhali | Local incidents and country state | আল্লার রমতে বেডিরা অনিম মেম্বরের মাইয়া। আল্লায় দিলে বাইরা দেইন্না অইছে। আঁর তো দেইন্না কেউ নাই হে অর্থে হেয়াল্লাইয়ে তো কইতেছি আমনে খোঁজ নি কতা কইয়েন সরকারেরা গর দিছে। ভাঙ্গা গর রাখি, কদিন বরা গর ঠিক কইত্তাইরবো অর্থ আছে। কিন্তুং শতানি করি গর গর দেনো। তো হেয়ারহরে সরকারে | By the grace of Allah, the women are the daughters of Anim Member. Since Allah has blessed the brothers, they are now able to give help. But I don't have anyone to give me anything in that sense. That's why I say you should check the factsrealize that after adding up the cost of seeds, before you speak. The government has provided houses, and they even have funds to repair damaged ones and make them livable within a few days. But out of evil intentions, they didn't give the house. And after that, the government. |
| train_rangpur_0222.wav | Rangpur | Childhood and old memories | মাইনষের আম চুরি করসি। কত কী করসি! তোর বন্ধু-বান্ধব সহকারে করসিল নাকি একলায়? বন্ধু-বান্ধব সহকারে করসিল। এখন তো বন্ধু-বান্ধব ও নাই। আগের মতো মজাও নাই। | I used to steal mangoes from people did all sorts of things! Did you do it with your friends or alone? I did it with my friends. Now there are no friends anymore, so it's not as fun as before. |
| train_sandwip_1044.wav | Sandwip | Food and Recipe | এডিক্কা বেড়াইন্না রাইনলেও এক মজা ভাই তরকারি। হ্যা, স্বাদ আছে। দুগা-দুগা দিয়ালাইছি আঁই কিরমু ক খান, আঁর হরি আলু টোয়া, আলুও দু-গা লগে দিছি,মাছ দুই ভাগ কইরলে তরকারি ভালা লাইগদ ন। হিয়াত্তনো দি দিছি। মা, মনি মোটর বন্ধ করি দ গোই। | Even when you cook it in a mixed-up or random way like this, the curry turns out really delicious. Yes, it has a nice taste. What can I do? My mother-in-law likes potatoes, so I added a couple of potatoes too. If I cut the fish into two pieces, the curry wouldn't taste as good. So I just put it in whole. Mamoni, please turn off the motor |
| train_sylhet_0139.wav | Sylhet | Leisure and Tours | খয়েকটা জেলা গুরছি যেমন চিটাগাং জেলা খুব বেশি গুরছি। খারণ চিটাগাং ইউনিভার্সিটিত ছিলান আমি আডারো দিন। ওই ইতার লাগি খুব গুরাগুরি খরছি। সবদিক দিয়া মিলাইয়া বাংলাদেশের প্রায় রাজশাহী জেলা গুরছি আর আমার দক্ষিণ <> আর মায়মানসিং এলাখার দিকে এখনো যাওয়া ওইছে না। | I have traveled to a few districts. For example, I have spent quite some time in the Chittagong district because I stayed at Chittagong University for eighteen days. That is why I traveled around the area a lot. All things considered, I have visited almost all parts of Bangladesh. I have been to the Rajshahi district and the southern regions, but I haven't gone to the Mymensingh area yet. |
| train_kishoreganj_0016.wav | Kishoreganj | Science and Technology | মানিক ভাইয়া। মোবাইল কিনসুইন বোলে? হ, ভাই। মোবাইল তো একটা লইছিলাম। লইছলাম ভাই। লইছলাইন। আচ্ছা, কি মোবাইল লইছুন। লইছিলাম ভাই, ওই যে রিয়েলমি চি ওয়ান ফাইভ। রিয়েল মি চি ক ওয়ান ফাইভ আচ্ছা। | Manik bhaiya, did you buy a mobile phone? Yes, brother, I bought one. Yes, I did — you bought one too, right? Alright, what mobile did you buy? I bought that one — the Realme C15. Realme C15, okay. |
| train_tangail_0910.wav | Tangail | Miscellaneous | কয় হাজতে দিমু। আমি শুইনাই তাজ্জব কারণ তোর বাপ বাইরই হইবার দেয় না মেয়ে মানুষরে বাসা থিকা। তাইলে নিজেরডা দিবো ক্যা? হুম। নিজের মেয়া জাত করবার পরে নাই বুঝছো? হুম, হুম। বাড়ি-ঘরে সবটি েয়া শাসনে রাইখপার পারছে কিন্তু | He said he would allow her to dress up. I was quite surprised to hear that, since your father doesn't even permit other women to leave the house. So why would he allow his own daughter to do so? It seems he couldn't discipline his own child, you see. Yet he managed to keep all the other women in the household under strict control. |
| train_habiganj_0800.wav | Habiganj | Religions and Festivals | ওই মুসলিম, মুসলমানদের ওইছে গিয়া দুইটা ইদ। কুরবানি ইদ আর রুজার ইদ। রুজার ইদ থিকা কুরবানি ইদ সব চেয়ে বালা লাগে। আর ইন্দুদের তো ধর্মীয় পূজা | There are two Eid holidays for Muslims. One of them is Eid-ul-Fitr and the other is Eid-ul-Adha. Eid-ul-Adha is the most enjoyable holiday. Hindus also have their own religious holidays. |

Table 5: Topic-specific examples from the corpus (2)

| Districts | Subregions | Count |
|---|---|---|
| Rangpur | Kamarpara | 1 |
| Kishoreganj | Pakundia, Bhairab, Bogadiya | 3 |
| Narail | Barakalia, Chotta Kalia, Kartikpur, Mirzapur, Ramnagar, Joka, Baka, Patna, Chandpur, Uthali, Jogania, Kalabaria, Pahardanga, Bil Bauchh, Joypur, Baidhyarbati, Gachhbaria | 17 |
| Chittagong | Nalanda, Satkania, Patharghata, Rangunia, Potiya, Hathazari, Boalkhali | 7 |
| Narsingdi | Polashtoli, Madhabdi | 2 |
| Tangail | Hemnagar, Shimlapara, Bholarpara | 3 |
| Habiganj | Bamkandi, Montoil, Putia, Baniachong, Chunarughat | 5 |
| Barishal | Jongolpotti | 1 |
| Sylhet | Tilargaon, Naya Bazar, Akhaliya, Kanaighat, Gowainghat, Dayamir, Tuker Bazar, Ambarkhana, Bianibazar, Bandar Bazar, Boroi Kandi, Chawkidekhi, Dargah Gate, Dariapara, Housing Estate, Jalalabad, Zindabazar, Kanishail, Kalapara, Kumarpura, Landani Road, Madina Market, Majumdari, Nawab Road, Pathantula, Roy Nagar, Sagar Dighir Par, Baluchar, Tilagarh, Fenchuganj, Golapganj, Baotila, Jaintia, Shahparan, Goalabazar, Chatul, Bagha, Hatimganj, Jaintapur, Bishwanath, Ghasitula, Companiganj, Kanaighat, Shibganj, Jaflong, Bhadeswar | 46 |
| Sandwip | Gachhua | 1 |
| Noakhali | Kobirhat, Shonapur | 2 |
| Comilla | Unjhuti, Borarchar, Sripur, Chhaliyakandi, Mohammadpur, Uthkhara, Elahabad, Gunaighar, Bangora, Chhepara, Barur | 11 |
| **Total** | - | **99** |

Table 6: Subregions covered from each district

| Districts | Sample Counts | | | Duration [H:M] | | | OOD% | | | Gender | | | Words Per Minute |
|---|---|---|---|---|---|---|---|---|---|---|---|---|---|
| | Train | Valid | Test | Train | Valid | Test | Train | Valid | Test | M | F | B | |
| Rangpur | | 1,298 | | | 6:00 | | | 51.93 | | | 27 | | 134.42 |
| | 1,038 | 130 | 130 | 4:48 | 0:36 | 0:35 | 50.60 | 36.22 | 45.17 | 19 | 8 | 0 | |
| Kishoreganj | | 2,049 | | | 9:35 | | | 55.71 | | | 47 | | 118.78 |
| | 1,639 | 205 | 205 | 7:42 | 0:55 | 0:58 | 62.91 | 55.49 | 48.08 | 25 | 18 | 4 | |
| Narail | | 1,859 | | | 8:36 | | | 57.23 | | | 37 | | 136.86 |
| | 1,487 | 186 | 186 | 6:52 | 0:53 | 0:51 | 56.49 | 43.35 | 44.79 | 21 | 12 | 4 | |
| Chittagong | | 1,757 | | | 8:11 | | | 63.83 | | | 41 | | 134.58 |
| | 1,405 | 176 | 176 | 6:35 | 0:48 | 0:47 | 62.07 | 57.47 | 64.63 | 15 | 22 | 4 | |
| Narsingdi | | 1,373 | | | 6:20 | | | 54.74 | | | 26 | | 148.53 |
| | 1,099 | 137 | 137 | 5:03 | 0:39 | 0:37 | 53.84 | 40.80 | 39.79 | 9 | 16 | 1 | |
| Tangail | | 1,271 | | | 6:12 | | | 32.36 | | | 36 | | 146.52 |
| | 1,017 | 127 | 127 | 5:03 | 0:35 | 0:34 | 45.13 | 27.02 | 24.74 | 18 | 11 | 7 | |
| Habiganj | | 1,170 | | | 5:26 | | | 60.82 | | | 34 | | 123.47 |
| | 936 | 117 | 117 | 4:19 | 0:32 | 0:34 | 58.06 | 59.34 | 56.65 | 19 | 15 | 0 | |
| Barishal | | 1,006 | | | 4:45 | | | 53.54 | | | 26 | | 124.57 |
| | 804 | 101 | 101 | 3:47 | 0:29 | 0:29 | 50.65 | 47.24 | 48.30 | 6 | 7 | 13 | |
| Sylhet | | 7,624 | | | 36:15 | | | 67.75 | | | 94 | | 128.01 |
| | 6,100 | 762 | 762 | 28:51 | 3:42 | 3:42 | 66.68 | 60.49 | 58.19 | 62 | 30 | 2 | |
| Sandwip | | 1,310 | | | 6:02 | | | 63.52 | | | 26 | | 144.26 |
| | 1,048 | 131 | 131 | 4:48 | 0:37 | 0:37 | 62.69 | 53.48 | 53.48 | 15 | 9 | 2 | |
| Comilla | | 318 | | | 1:27 | | | 51.40 | | | 5 | | 166.26 |
| | 254 | 32 | 32 | 1:09 | 0:08 | 0:08 | 55.16 | 23.26 | 21.77 | 3 | 2 | 0 | |
| Noakhali | | 278 | | | 1:16 | | | 49.02 | | | 7 | | 131.95 |
| | 222 | 28 | 28 | 1:00 | 0:07 | 0:08 | 48.37 | 40.36 | 43.24 | 3 | 4 | 0 | |
| **TOTAL** | | 21,313 | | | 100:15 | | | 76.05 | | | 394 | | 135.62 |
| | 17,049 | 2,132 | 2,132 | 80:03 | 10:06 | 10:05 | 74.82 | 62.89 | 62.45 | 209 | 148 | 37 | |

Table 7: Regional Speech Corpus Statistics: OOD = words unique to the region that are Out Of Dictionary in comparison to SCB. | Gender (M-F-B) = M → Male, F → Female, B → Both | H:M = Hour(s) : Minute(s)